\documentclass[a4paper, 10pt, conference]{IEEEtran}
\IEEEoverridecommandlockouts  
\input epsf
\usepackage{url}
\usepackage{comment}
\usepackage{makecell}
\usepackage{graphicx}
\usepackage{graphicx,subcaption,lipsum}
\usepackage{algorithm}
\usepackage{algorithmic}
\usepackage{soul}
\usepackage{multirow}
\usepackage{csquotes}
\usepackage[english]{babel}
\usepackage{amsthm}
\usepackage{amssymb}
\usepackage{subcaption}
\usepackage{url}
\usepackage{amsmath}
\usepackage{xcolor}   
\usepackage{amssymb}
\usepackage[utf8]{inputenc}
\usepackage{geometry}
\usepackage{tikz}
\usepackage{cite}
\usetikzlibrary{shapes, arrows, positioning}
\DeclareRobustCommand*{\IEEEauthorrefmark}[1]{%
\raisebox{0pt}[0pt][0pt]{\textsuperscript{\footnotesize\ensuremath{#1}}}}
\begin{document} 

\title{\LARGE \bf Dynamic Hand Gesture Recognition for Robot Manipulator Tasks}
\author{\IEEEauthorblockN{Anonymous Authors}}
\author{\IEEEauthorblockN{Dharmendra Sharma\IEEEauthorrefmark{},
Peeyush Thakur\IEEEauthorrefmark{}, Sandeep Gupta, Narendra Kumar Dhar\IEEEauthorrefmark{}, and Laxmidhar Behera} 
\IEEEauthorblockA{Indian Institute of Technology Mandi, India}}
\maketitle
\IEEEpubidadjcol
\begin{abstract}
This paper proposes a novel approach to recognizing dynamic hand gestures facilitating seamless interaction between humans and robots. Here, each robot manipulator task is assigned a specific gesture. There may be several such tasks, hence, several gestures. These gestures may be prone to several dynamic variations. All such variations for different gestures shown to the robot are accurately recognized in real-time using the proposed unsupervised model based on the Gaussian Mixture model. The accuracy during training and real-time testing prove the efficacy of this methodology.

\end{abstract}
\pagenumbering{gobble}

\section{\textbf{Introduction}}
Modern industries such as manufacturing, agriculture, entertainment, security, healthcare, food preparation, military, and customer services depend heavily on robotics, which calls for tedious, repetitive labor demanding extreme precision. Robots can be used in these sectors along with humans since they are precise, trained to be clever, and never get tired like humans do \cite{1}. Many of the tasks require human interaction in making decisions. One such interaction may be through gestures.

The term \enquote{gesture} refers to the physical movements of the hand, body, head, or face that convey an opinion, emotion, or other information. Hand gestures are very common in different tasks. Recognition of such gestures involves real-time tracking and interpretation of movements made by a person's hands. With sensors (cameras or depth-sensing devices), this technology captures dynamic gestures, recognizing specific patterns or configurations to infer intended actions. Hand gestures have applications in various fields, such as robotics, virtual reality, gaming, and human-computer interaction. They enable users to communicate with or interact with devices offering a more natural and immersive interface without requiring physical touch or traditional input devices \cite{2}\cite{3}.
This paper presents an unsupervised learning approach based on the Gaussian mixture model (GMM). It is widely used in multiple fields such as robotics and autonomous systems, image and video processing, natural language processing, anomaly detection, medical imaging, etc.\cite{5}.
It is assumed that data points are generated from a combination of multiple Gaussian distributions. The primary objective is to estimate the precise parameters of these Gaussians, along with the probabilities of each data point belonging to each Gaussian component. GMM permits data points to be part of several clusters with respective probabilities. This approach works well in handling dynamic data that has overlapping categories \cite{4}. The hand gestures, here, are used to command the robot manipulator to perform different tasks. Each gesture represents a specific task. It is very important to identify the gestures correctly for task execution.

\begin{figure}[h!]
    \centering
    \begin{subfigure}{0.47\textwidth} 
        \centering
        \includegraphics[width=1\linewidth]{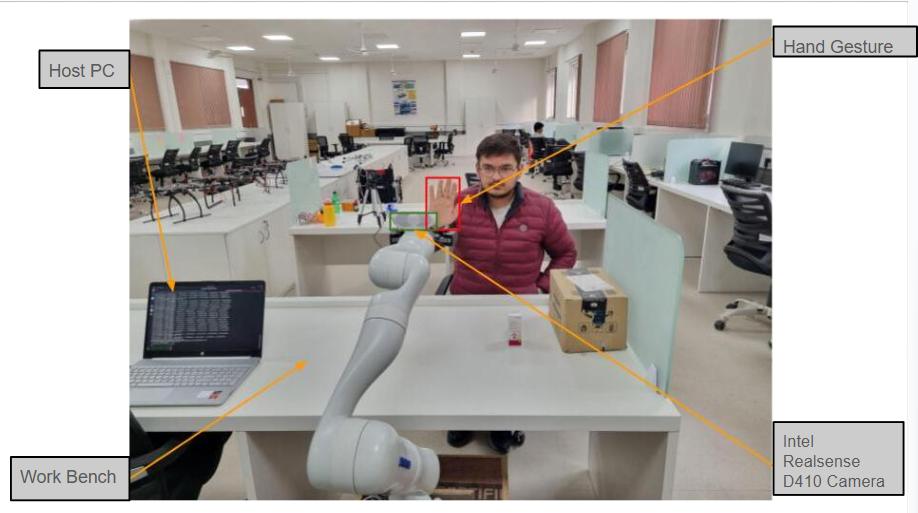}
    \end{subfigure}
    \caption{Real-time hand gesture recognition using an RGB-D camera attached to the robot manipulator. }
    \label{fig:1}
\end{figure}
The \textbf{ practical motivation} for this paper is to address some of the issues in gesture recognition during such tasks: i) \textit{precision and accuracy:} achieving high precision and accuracy while capturing intricate hand movements; ii) \textit{variability and context sensitivity:} addressing the variability among individuals and the context sensitivity of gestures are crucial for reliable identification; and iii) \textit{real-time processing:} developing efficient real-time processing capabilities is essential to ensure quick and accurate interpretation of gestures, especially in dynamic environments. The model proposed in this paper takes care of the above aspects.

The paper is organized as follows: Section II presents the literature review and problem formulation; Section III presents a proposed methodology; Section IV presents a convergence of the proposed methodology; Section V presents results and discussion; and finally, Section VI concludes the paper.

\section{\textbf{Literature Review and Problem formulation}}
Recognizing gestures can be quite challenging because they involve different factors, such as understanding body movement, analyzing the way they are made, recognizing patterns, and using machines to learn and understand the movements.

Several contributions, ranging from fundamental work \cite{rashid12}, \cite{6} to recent ones \cite{7}, \cite{8}, \cite{song21}, \cite{bandara23} and their references collectively emphasize the significant strides in vision-based hand gesture recognition and the need for continued research to propel human-robot interaction on natural terms. They emphasize the evolving landscape of vision-based hand gesture recognition systems, highlighting advancements in feature extraction, classification algorithms, and gesture representation. Innovative adaptations such as the equal-variance GMM for image classification showcase potential improvements in image representation techniques. The integration of methodologies from the speech and image domains offers a valuable cross-disciplinary perspective. 

Boruah \textit{et. al.} \cite{9} address the escalating issue of falls among the elderly by presenting an innovative intelligent fall detection method named as pose estimation-based fall detection methodology (PEFDM). Operating on an edge-computing architecture, PEFDM utilizes human body posture recognition to efficiently detect falls without compromising privacy or requiring extensive image transfer. This dual-pronged approach showcases the versatility of hand gesture recognition systems.

The work in \cite{10} proposes an interactive learning aid tool demonstrating its effectiveness in controlling various virtual elements by recognizing six distinct hand gestures. Hyder \textit{et. al.} \cite{11} introduce a vision-based hand gesture recognition system designed to enhance communication and interaction in e-learning environments. MediaPipe is used for hand gesture recognition and a virtual-mouse-based object control system for navigating virtual objects in Unity, improving user experience through intuitive control of virtual objects.

The work in \cite{12} focuses on contactless gesture recognition, addressing challenges like diverse palm sizes and occlusions using deep neural networks. MediaPipe aids in hand landmark detection, and four neural network types (RNN, CNN, hybrid, and Transformer Encoder) identify ten gestures (0 to 10). The study envisions practical applications in contactless human-computer interaction. Gupta \textit{et al.} \cite{13} compares VGG16, InceptionNet, EfficientNet, and a self-designed CNN for hand gesture recognition. The proposed CNN model, with minimal trainable parameters, achieves high accuracy, making it suitable for edge device deployment. In addition, Jiang \textit{et al.} \cite{du18} used depth information with CNN for hand gestures recognition on ASL dataset.

The approaches developed in the above works are useful in several scenarios but have limitations such as dynamic variations in gestures, background disturbances, etc. The work proposed in this paper does not have these limitations.

\subsection{Problem formulation}
The challenge is to develop a real-time solution that can identify hand gestures accurately for robot manipulator tasks. Current methods struggle with precision and responsiveness, especially in dynamic environments. The goal is to create a reliable model that swiftly interprets diverse hand gestures and translates them into precise commands for robot manipulator tasks. The key features of this model include accommodating gesture pattern variability, adapting to dynamic surroundings, and minimizing latency for quick and accurate robot responses. Silhouette score, a performance metric that verifies the model's efficiency in real-time human-robot interaction.

\section{\textbf{Proposed Methodology}} 
The proposed methodology has three parts described in the following subsections.
\subsection{System Setup}
A 7-DOF (Kinova Gen3) robot manipulator is used in this work for executing different tasks. A visual sensor (i.e., camera) on its top captures the video of hand gestures in real time. There is no restriction on background. The setup comprising the RGB-D camera on the robot manipulator is shown in Fig. \ref{fig:1}.
\subsection{Hand Gesture Recording and Initial Processing}
The following steps are considered once the video is available for initial processing.
\begin{itemize}
\item The raw gesture data needs to be pre-processed to enhance the quality and relevance of the information from the inherent features. We must select dominant ones from several features that can depict any task. The pre-processing involves normalization, noise reduction, and feature extraction to transform raw data into a suitable format for training the identification model (refer to Section III-C).
\item The pre-processed video is divided into a certain number of image frames.
\begin{figure}[htp]
\centering
\subfloat[] {\includegraphics[width=.24\linewidth]{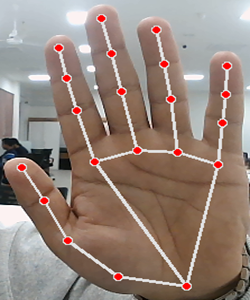}}
\subfloat[] {\includegraphics[width=.24\linewidth]{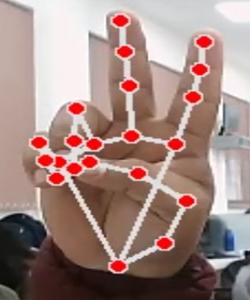}}
\subfloat[] {\includegraphics[width=.24\linewidth]{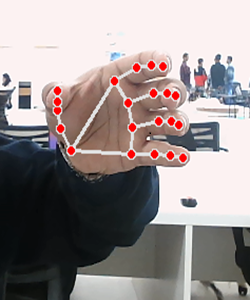}}
\subfloat[]{\includegraphics[width=.24\linewidth]{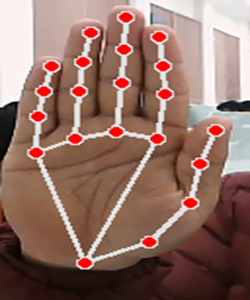}}
\caption{Hand gestures for different tasks: (a) wave, (b) pick, (c) stack, and (d) push.}
\label{fig:2}
\end{figure}
\item The image frames also contain background, hence, to remove entities other than the hand, the region-based segmentation algorithm is used.
\item These segmented frames are processed using the histogram of gradient (HOG3D) algorithm to extract the features needed to find the pattern in gestures. It also reduces the dimensionality of the raw data by removing unnecessary information such as noise and lighting effects, which enhances computational efficiency.
\item The extracted features provide the location of the landmark points (i.e., points on the palm and each finger), thereby, movement of the hand in the $x$, $y$, and $z$ directions. There are $21$ such landmark points as shown in Fig. \ref{fig:2}.
\end{itemize}
The gestures specific to tasks are then identified based on the relative position of the landmark points.
\subsection{Gesture Identification} 
This approach leverages the strengths of GMMs to discern distinct patterns within gesture data. The GMM training is applied to features based on landmark points derived from all the frames in a video. Once complete, the final decision on gesture for task type is made. The key steps are discussed below.
\subsubsection{Features from each Video}  
Once the initial processing is complete (Section III-B), compute the variance of each landmark point for all the frames to determine how much each point varies in the entire video. Hence, from each video, a feature matrix of $21 \times 3$ dimension is obtained with $x$, $y$, and $z$ variances for $21$  landmark points. To train the proposed model, real-time videos of different gestures are used.
\subsubsection{GMM Training on Features}
Once the feature vectors are prepared, the model is trained. 

This model involves probability distribution as a weighted sum of Gaussian distributions \cite{16}, each characterized by its mean, covariance, and mixing weight coefficient based on the extracted features from the gestures to recognize patterns in them. The number of Gaussian distributions is initialized based on the number of tasks. Each gesture is represented by the probability density function
\begin{align}\label{eq1} 
&p(X)=\sum_{k=1}^{K} \pi_k p_k(X).\\
\text{with}\quad &0 \leq \pi_k \leq 1 \,\,\,\text{and}\,\,\,\sum_{k=1}^K \pi_k=1, \notag
\end{align} 
where $X$ is the dataset comprising gestures for $K$ different tasks with dynamic variations of each gesture. $\pi_k$ is the mixing weight coefficient, representing how well each gesture type reflects in a particular gesture.
\par 
The components $p_k(X)$ are from normal distributions. Such a distribution is considered to model the underlying patterns in the continuous real-time gesture data. Other basic distributions (e.g., Bernoullis, Gammas, and Gaussians) may be considered depending on the nature of the data.

The movements in a gesture may exhibit more complex patterns that may not be modeled accurately by a simple Gaussian distribution, hence, GMM is used.
The probability distribution is computed to check whether data points belong to the correct gesture. 

The probability of any landmark point in a given gesture frame is equal to the conditional probability of dataset point $X$ for a given  parameter $z$, i.e.,
 \begin{equation}\label{eq2}
p({X}\mid {z}) = \mathcal{N}\left({X} \mid {\mu},{\Sigma}\right).
\end{equation} 

GMM being a parametric model represents the gesture data as a combination of multiple univariate normal distribution (i.e., multivariate distribution) components, where each component represents a cluster in the data. 
Hence, (\ref{eq2}) can be written as
\begin{align}\label{eq4}
p\left(X \mid \mu, \sigma^2\right)=&\mathcal{N}\left(\mu, \sigma^2\right)=\frac{1}{(2 \pi)^{d/2} {|\Sigma|}^{1/2}}\times\notag \\ &\exp \left(-\frac{1}{2}(X-\mu)^{\top} \Sigma^{-1}(X-\mu)\right),
\end{align}
where $d$, $\mu$, $\sigma$, and $\Sigma$ represent the dimension of the gesture data set $X$, the mean of different normal distributions, standard deviation, and the co-variance within different normal distributions, respectively.
For a single task, the probability distribution of different gesture data $p_k(X)$ can be replaced with different normal distributions, hence, (\ref{eq1}) becomes
\begin{equation}\label{eq5}
p({X} \mid {\theta})=\sum_{k=1}^K \pi_k \mathcal{N}\left({X} \mid {\mu}_k,{\Sigma}_k\right),
\end{equation}
where $\theta$ represents the model parameters mean $(\mu)$, variance $(\Sigma)$, and mixing weight coefficients $(\pi_k)$ for every component. In \eqref{eq5}, there is a single gesture video for each task in the data set ${X}$. However, if there are multiple videos for each gesture with a total of $N$ videos then 
\eqref{eq5} is modified to
\begin{equation}\label{eq6}
p({X} \mid {\theta})=\prod_{n=1}^{N} \sum_{k=1}^K \pi_k \mathcal{N}\left({X_n} \mid {\mu}_k,{\Sigma}_k\right).
\end{equation}
The equation (\ref{eq6}) is the likelihood function for gesture distributions of different tasks. For ease of computation, log-likelihood $(L)$ is used in \eqref{eq6} given as
\begin{equation}\label{eq7} 
\textit{L} =\log p({X} \mid {\theta})=\sum_{n=1}^{N} \log \Big(\sum_{k=1}^K \pi_k \mathcal{N}\left({X_n} \mid {\mu}_k,{\Sigma}_k\right)\Big).
\end{equation} 
The objective is to find the maximum probability for a given gesture, thereby, a need to compute the maximum of the log-likelihood in \eqref{eq7}. This is done by updating the model parameters in \eqref{eq7} until convergence occurs. For this, the \textit{expectation-maximization} algorithm is used. The probability of every landmark point for a gesture in each video is computed and is represented as the elements of \textit{responsibility matrix} \cite{14} in Table $I$. 
\begin{table}[]
\caption{Probability of landmark points ($l_1, ..., l_{21}$) in each image frame of a gesture video ($G_1, G_2, G_3, G_4$ are gestures)}
\footnotesize
\resizebox{0.95\linewidth}{!}{%
\begin{tabular}{c|c|c|c|c|}
\cline{2-5}
$l_1$                                               & $p_{1} (G_1)$  & $p_{1} (G_2)$  & $p_{1} (G_3)$  & $p_{1} (G_4)$  \\ \cline{2-5}
$l_2$                                                & $p_{2} (G_1)$  & $p_{2} (G_2)$  & $p_{2} (G_3)$  & $p_{2} (G_4)$  \\ \cline{2-5}
\begin{tabular}[c]{@{}c@{}}.\\ .\\ .\end{tabular} & ...           & ...           & ...           & ...           \\ \cline{2-5}
$l_{21}$                                            & $p_{21} (G_1)$ & $p_{21} (G_2)$ & $p_{21} (G_3)$ & $p_{21} (G_4)$ \\ \cline{2-5}

\end{tabular}%
}
\end{table}

The highest probability of a landmark point means it belongs to that particular gesture. The matrix is continuously updated with the probability values based on the updated model parameters till convergence occurs (refer Algorithm \ref{alg: EM}).

\begin{algorithm}\label{alg1}
\caption{Computation of model parameters}\label{alg: EM}
\begin{algorithmic}[1]
\REQUIRE Initially, the following parameters are considered:
 $K$: number of Gaussian distributions based on number of gestures\\ 
 $k$: an individual gesture;  $n$: number of landmark points\\
 $\mu_k$, $\Sigma_k$, $\pi_k$: mean, covariance, and mixing weight parameter for each distribution, respectively\\ 
\ENSURE~$K$ Number of Clusters\\
\textbf{begin:}
\WHILE{no convergence}
\STATE\textbf{Expectation} The responsibility matrix element $r_{nk}$ is the probability measure of a landmark point for a particular gesture. It is computed as
 \begin{equation}\notag
  r_{nk}=\frac{\pi_k\mathcal{N}(X_n\mid \mu_k, \Sigma_k)}{\sum_{j=1}^{K}\pi_j\mathcal{N}(X_n\mid \mu_j, \Sigma_j)}.
 \end{equation}
\STATE \textbf{Maximization}
\begin{itemize}
    \item \textit{Mean ($\mu_k$) update:} Maximize $L$ in \eqref{eq7} with respect to $\mu_k$ (i.e., $\frac{\delta L}{\delta \mu_k}=0$) to obtain 
    
 \begin{equation}\label{eq8}
  {\mu_k}^{new}=\frac{\sum_{n=1}^{N} r_{nk} X_n}{N_k}. 
 \end{equation}
 \item \textit{Co-variance ($\Sigma_k$) update:} Maximize $L$ in \eqref{eq7} with respect to $\Sigma_k$ (i.e., $\frac{\delta L}{\delta \Sigma_k}=0$) to obtain
\begin{equation}\label{eq9}
{\Sigma_k}^{new}=\frac{\sum_{n=1}^{N} r_{nk}(X_n-\mu_k) (X_n-\mu_k)^\top}{N_k}.
\end{equation}
\item \textit{Mixing weight coefficient ($\pi_k$) update:} The Lagrange multiplier ($\lambda$) is used with the $\log$-likelihood function in \eqref{eq7} as
\begin{equation}\label{eq10}
\mathcal{L}= L+ \lambda \sum_{k=1}^{K}\pi_k -1.
\end{equation}
On maximizing $\mathcal{L}$ with respect to $\pi_k$ (i.e., $\frac{\delta L}{\delta \Sigma_k}=0$), the updated $\pi_k$ is
\begin{equation}\label{eq12}
{\pi_k}^{new}=\frac{N_k}{N}.
\end{equation}
\end{itemize}
 \ENDWHILE \\
\textbf{end}
\end{algorithmic}
\end{algorithm}

\subsubsection{Gesture Assignment based on Features}
The new gesture instances are classified based on their maximum probability in the responsibility matrix. Table II showcases an example considering one video. Here, each landmark point across all the frames in the video has probabilities for all the $4$ gestures. The maximum probability shows its inclination towards a particular gesture (eg., $l_1$ shows towards $G_3$). 
An overall count for the $4$ gesture types is obtained for all the points. The gesture with the maximum overall count is assigned to the video data.
\begin{table}[h!]
\caption{Gesture assignment based on each landmark point.}
\footnotesize

\resizebox{\linewidth}{!}{%
\centering
\begin{tabular}{c|c|c|c|c|c}
\cline{2-5}
& $G_1$  & $G_2$  & $G_3$  & $G_4$  &                                                   \\ \cline{2-5}
$l_1$                                                & 0.2 & 0.1 & 0.6 & 0.1 & max($p$($l_1$)) = 0.6                                        \\ \cline{2-5}
$l_2$                                              & 0.1 & 0.5 & 0.2 & 0.2 & max($p$($l_2$)) = 0.5                                     \\ \cline{2-5}
\begin{tabular}[c]{@{}c@{}}.\\ .\\ .\end{tabular} & ... & ... & ... & ... & \begin{tabular}[c]{@{}c@{}}.\\ .\\ .\end{tabular} \\ \cline{2-5}
$l_{21}$                                               & 0.08 & 0.02 & 0.8 & 0.1 & max($p$($l_{21}$)) = 0.8                                    \\ \cline{2-5}
\end{tabular}%
}
\end{table}
\subsubsection{Evaluation Metrics}
To assess the performance of the proposed GMM-based gesture identification system, standard evaluation metrics are employed, such as the Silhouette score. It is a metric to calculate the goodness of a clustering technique. Its value ranges from $-1$ to $1$. A higher value indicates that a gesture is well-matched to its cluster and poorly matched to neighboring clusters.

The entire gesture identification methodology for robot task execution is discussed in Algorithm (\ref{alg: HGCIA}). 

\begin{algorithm}
\caption{Gesture Identification for Task Execution}\label{alg: HGCIA}
\textbf{Input:} Capture a real-time gesture video of 5 seconds.\\
\textbf{Output:} Task execution based on the identified gesture.\\
\textbf{Begin:}\\
$1.$ \textit{Feature extraction and reduction:} Calculate the variances ($\sigma_x,\sigma_y,\sigma_z$) of every landmark point in each image frame (Section II-C-1). \\
$2.$ \textit{Training:} The output of Step $1$ is used to train the model (Section II-C-2 and II-C-3).\\
$3.$ \textit{Testing:} Load the trained model for testing different real-time gestures. The variance is calculated for landmark points in the image frames of the video. The variance is normalized as per the training model and probabilities of that gesture having its likelihood with different gestures are computed. The decision is made based on the maximum probability for a particular gesture. \\
$4.$ \textit{Execution:} The robot manipulator performs the task associated with that particular gesture.\\
\textbf{End}
\end{algorithm}

\section{Convergence of Proposed Methodology}
The convergence of the proposed methodology ensures precise gesture identification to guarantee its applicability to any dataset. The convergence here is decided based on the model parameters.\\
\textbf{Theorem.} \textit{Given the video features extracted from its different frames, the gradual update of model parameters (mean, co-variance, and mixing weight coefficient) renders the convergence to accurate gesture distributions or clusters.}\\
\textit{Proof.} The complete proof is given in \enquote{Appendix}.

\section{\textbf{Results and Discussions}}
\subsection{Experimental Setup}
i) \textit{Hardware:} The Kinova Gen3 robotic arm ($7$-DOF) \cite{kinova_gen3_user_guide,kinova_ros} is used that has a visual sensor (Intel RealSense Depth Module D410)
to take real-time gesture videos. 
A GPU (GeForce GT-710, clock speed of 954 MHz and 32GB RAM) is used for computation.

ii) \textit{Software:} The Robot Operating System (ROS) is used for control and interaction. We have used a Kortex driver installed on the host computer. The algorithms are implemented via Python 3.10 in Visual Studio. 
\begin{figure}
    \centering
    \begin{subfigure}{0.2\textwidth}
        \centering
        \includegraphics[width=0.76\linewidth]{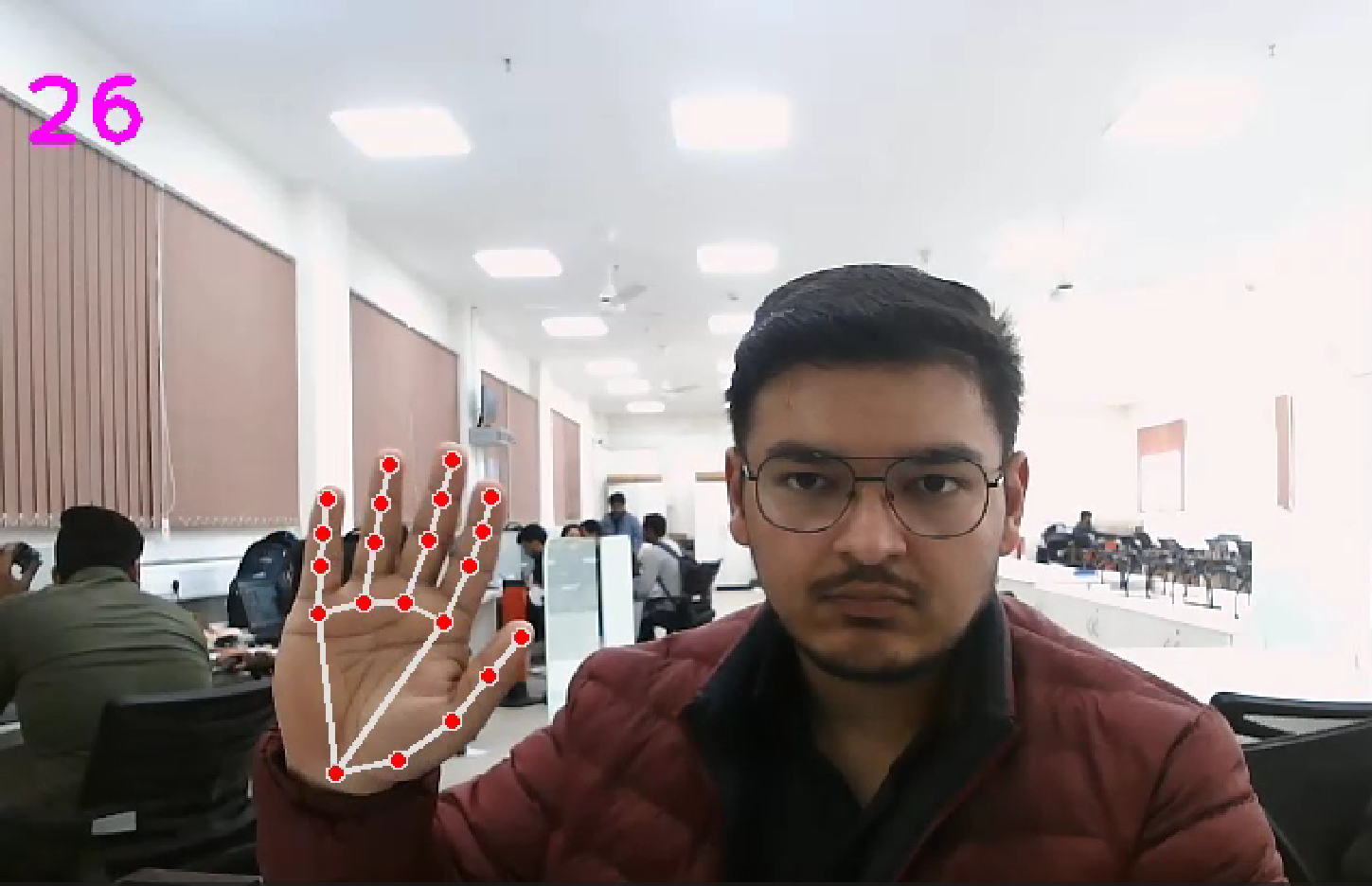}
        \caption{Wave.}
        \label{fig:sub1}
    \end{subfigure}
    \hspace{0.01\textwidth} 
    \begin{subfigure}{0.2\textwidth}
        \centering
        \includegraphics[width=1\linewidth]{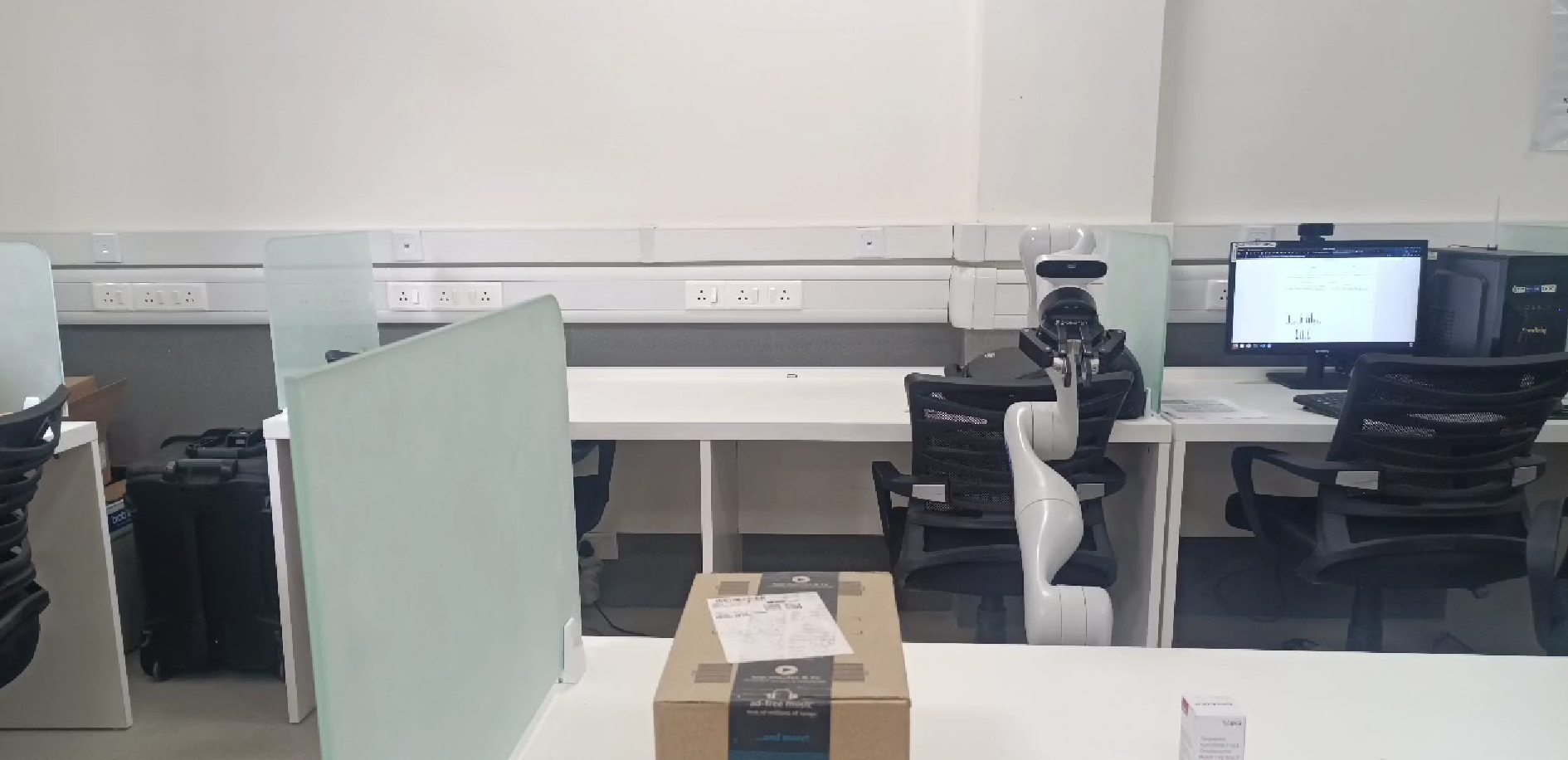}
        \caption{Robot Execution - Wave.}
        \label{fig:sub2}
    \end{subfigure}
    \hspace{0.01\textwidth} 
    \begin{subfigure}{0.2\textwidth}
        \centering
        \includegraphics[width=0.7\linewidth]{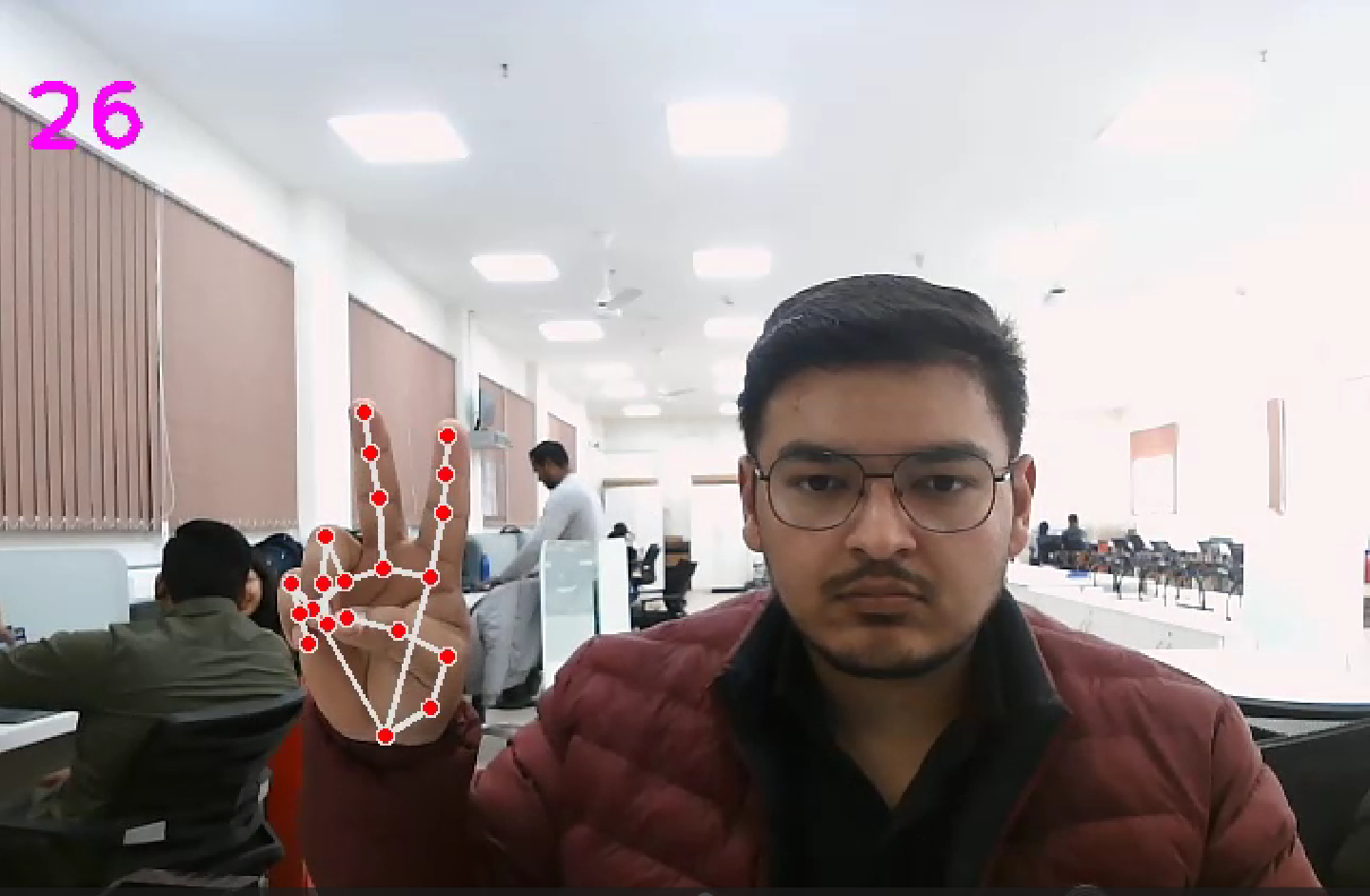}
        \caption{Pick.}
        \label{fig:sub3}
    \end{subfigure}
    \hspace{0.01\textwidth} 
    \begin{subfigure}{0.2\textwidth}
        \centering
        \includegraphics[width=1\linewidth]{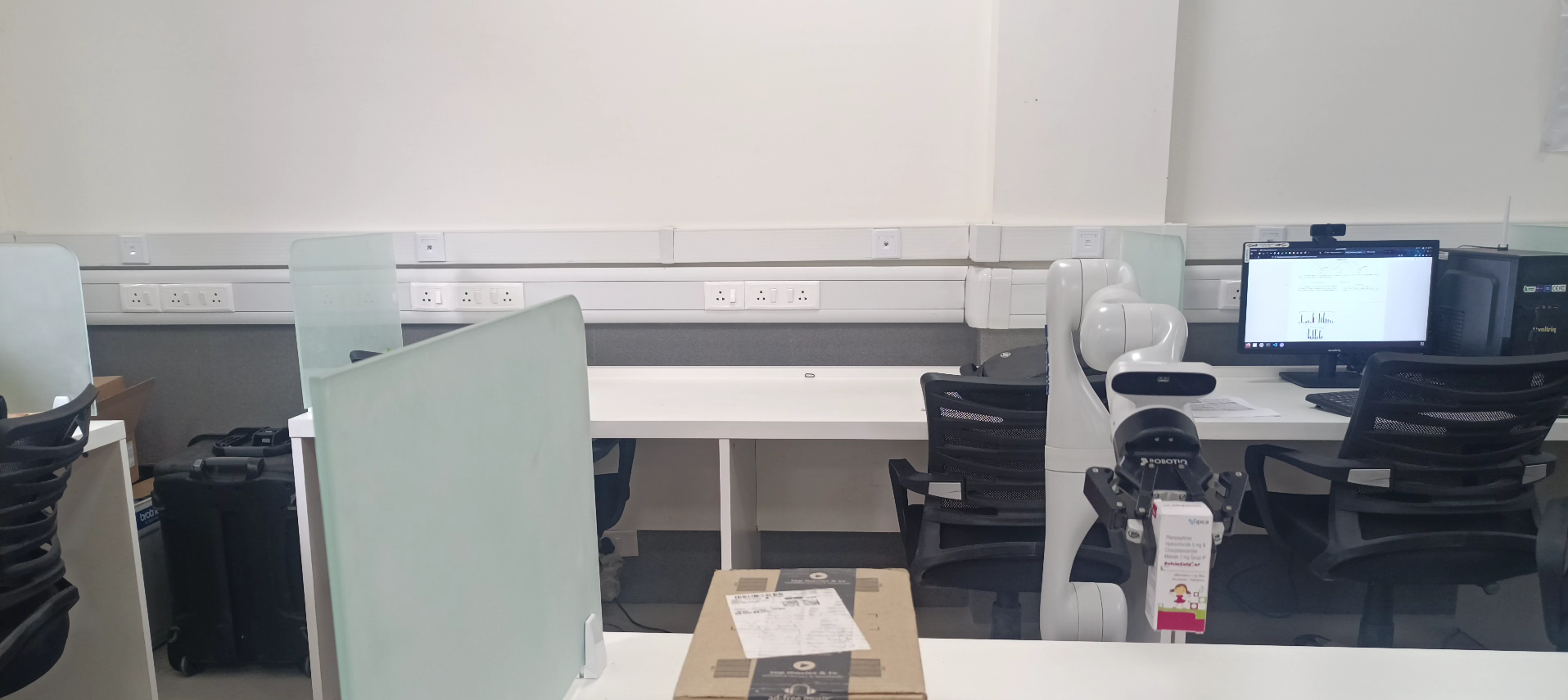}
        \caption{Robot Execution - Pick.}
        \label{fig:sub4}
    \end{subfigure}
    
    \vspace{0.01\textwidth} 
    
    \begin{subfigure}{0.2\textwidth}
        \centering
        \includegraphics[width=0.7\linewidth]{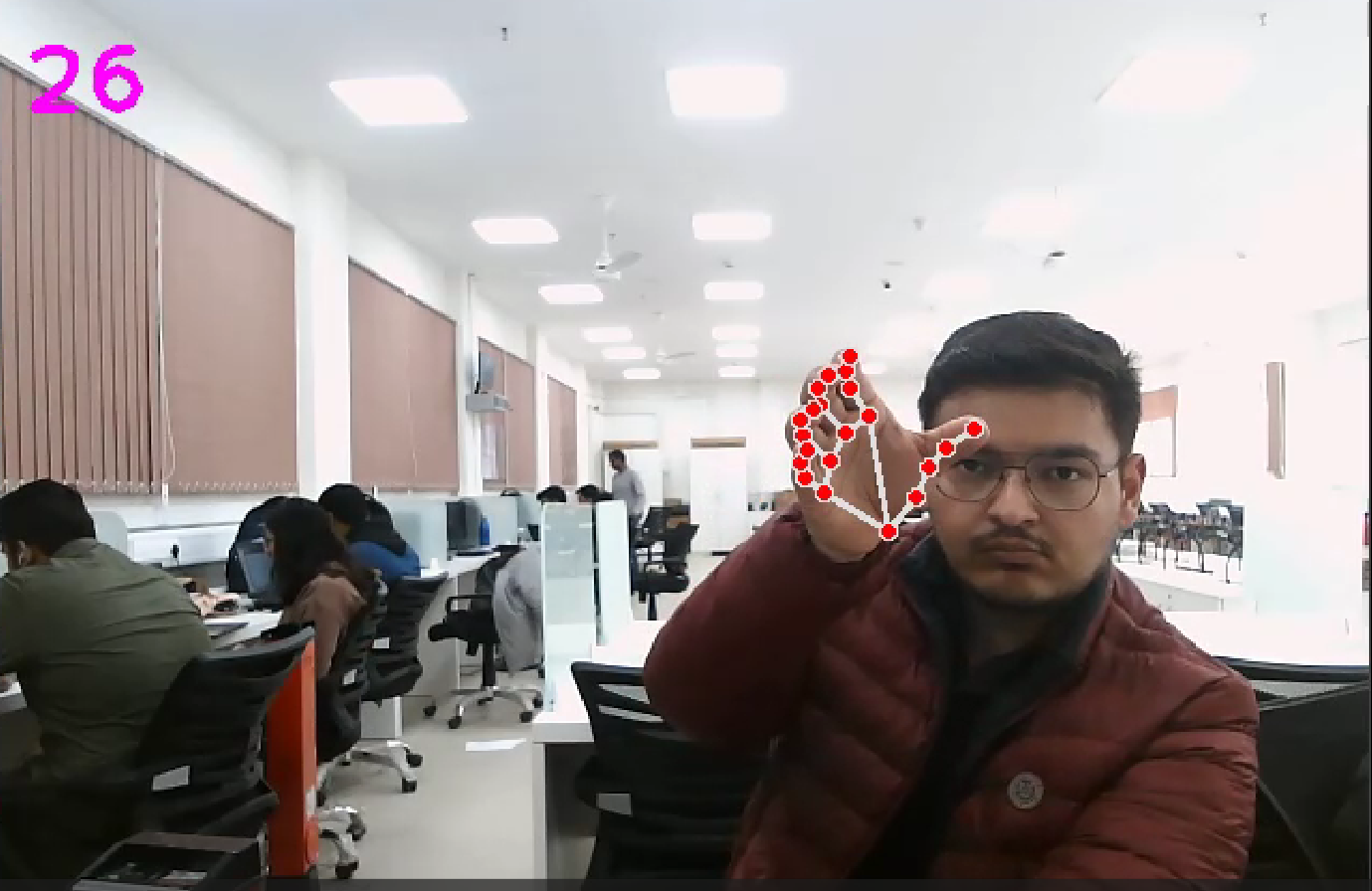}
        \caption{Stack.}
        \label{fig:sub5}
    \end{subfigure}
    \hspace{0.01\textwidth} 
    \begin{subfigure}{0.2\textwidth}
        \centering
        \includegraphics[width=1\linewidth]{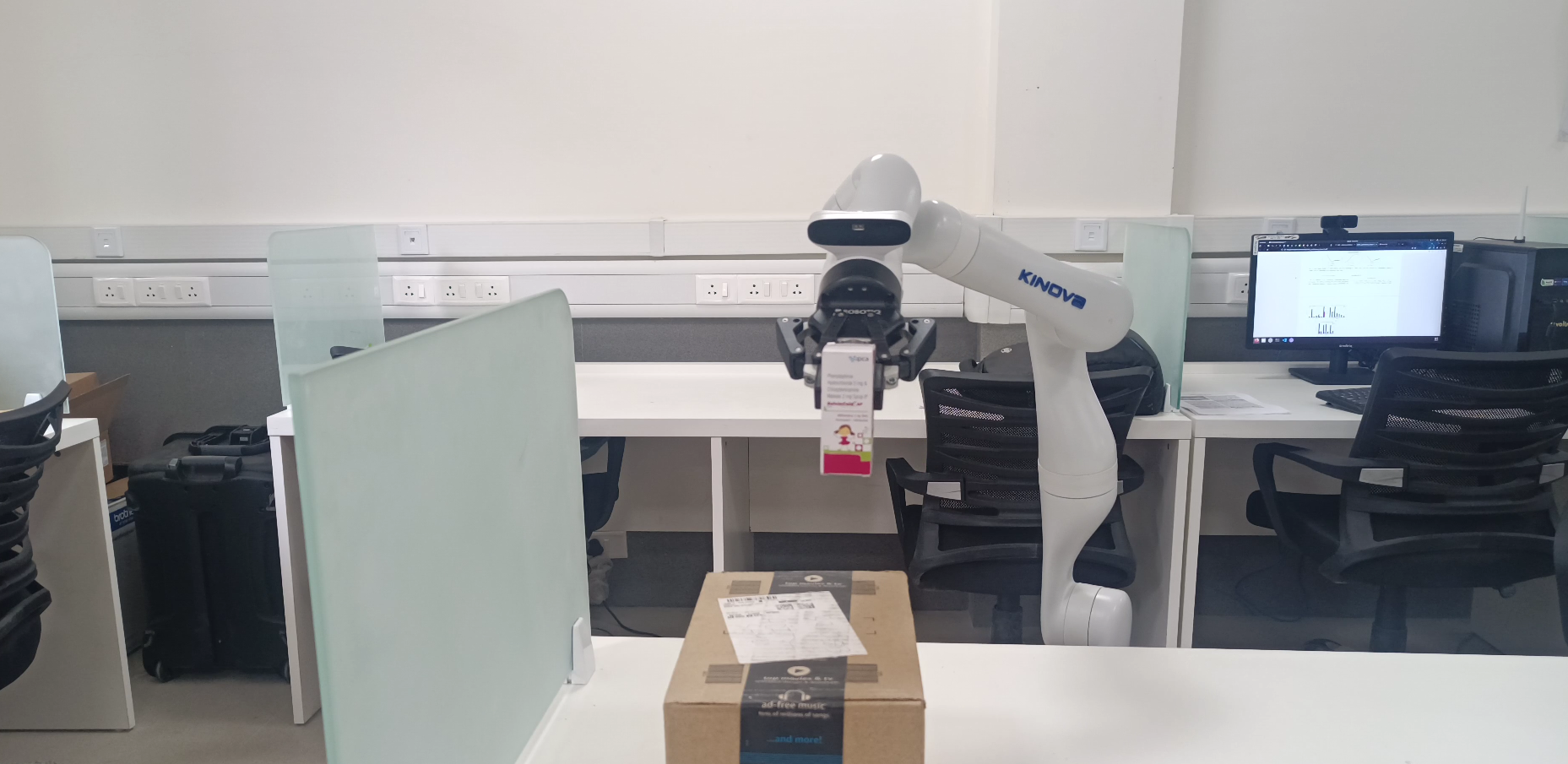}
        \caption{Robot Execution - Stack.}
        \label{fig:sub6}
    \end{subfigure}
    \hspace{0.01\textwidth} 
    \begin{subfigure}{0.2\textwidth}
        \centering
        \includegraphics[width=0.7\linewidth]{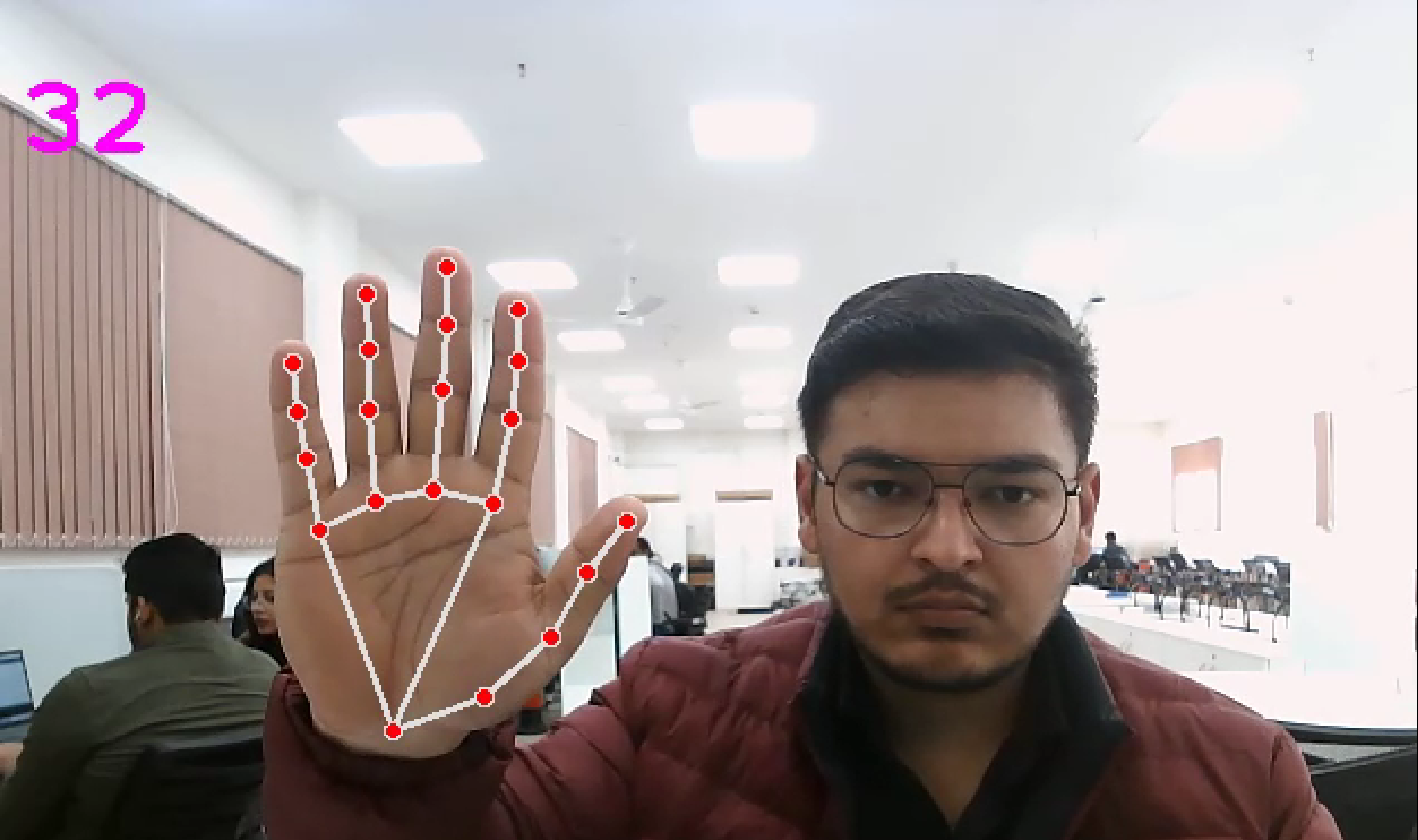}
        \caption{Push.}
        \label{fig:sub7}
    \end{subfigure}
    \hspace{0.01\textwidth} 
    \begin{subfigure}{0.2\textwidth}
        \centering
        \includegraphics[width=1\linewidth]{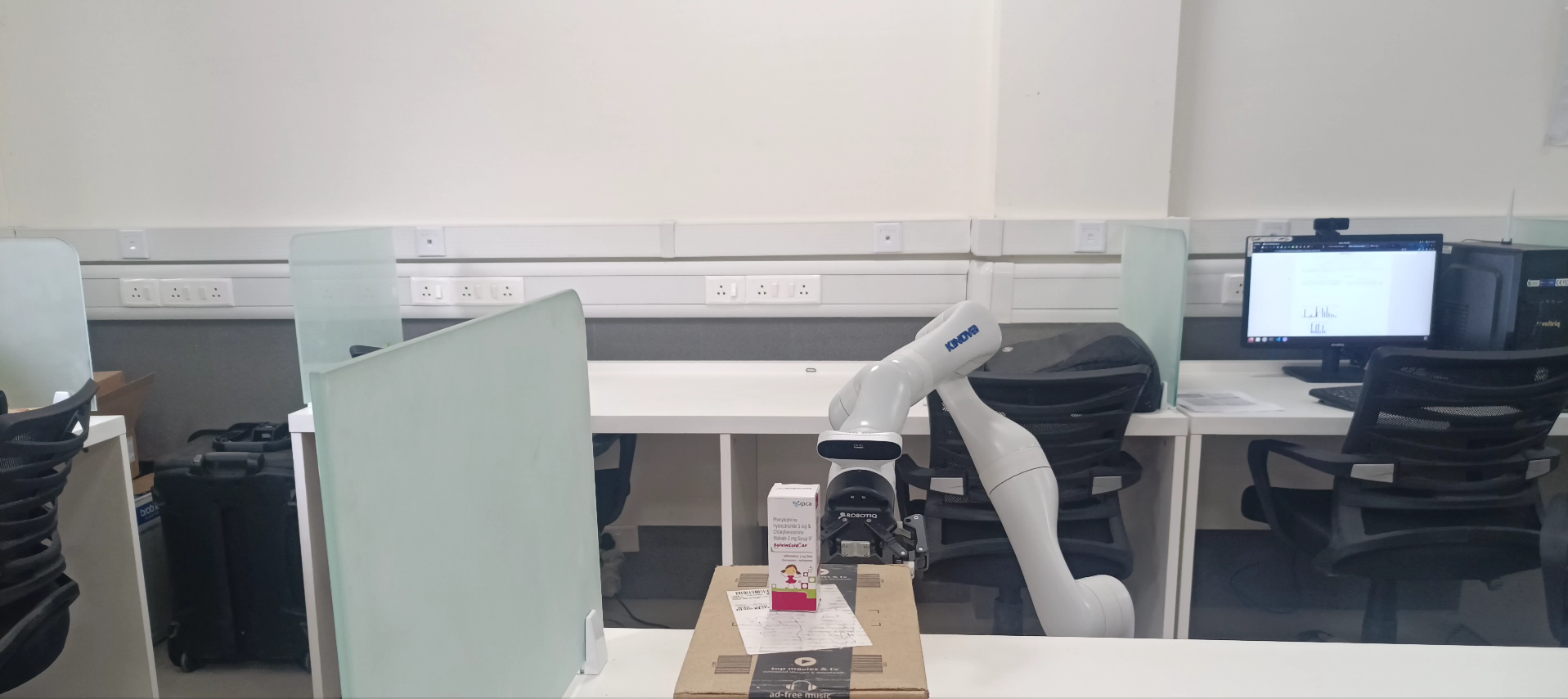}
        \caption{Robot Execution - Push.}
        \label{fig:sub8}
    \end{subfigure}

    \caption{Four dynamic hand gestures: (a) wave: movement largely in $x-y$ directions, (c) pick: movement largely in $y$ direction, (e) stack: movement in all $x-y-z$ directions, and (g) push: movement largely in $z$ direction. The robot tasks are: (b) initializing the robot gripper, (d) picking up an object, (f) stacking an object on a box, and (h) pushing the object. A video demonstrating all these tasks can be seen at \cite{youtube}.}
    \label{fig:test}
\end{figure}
 
\subsection{Experiments}

\begin{figure}[htp!]
\centering
\subfloat[Before training]{\includegraphics[width=.48\linewidth]{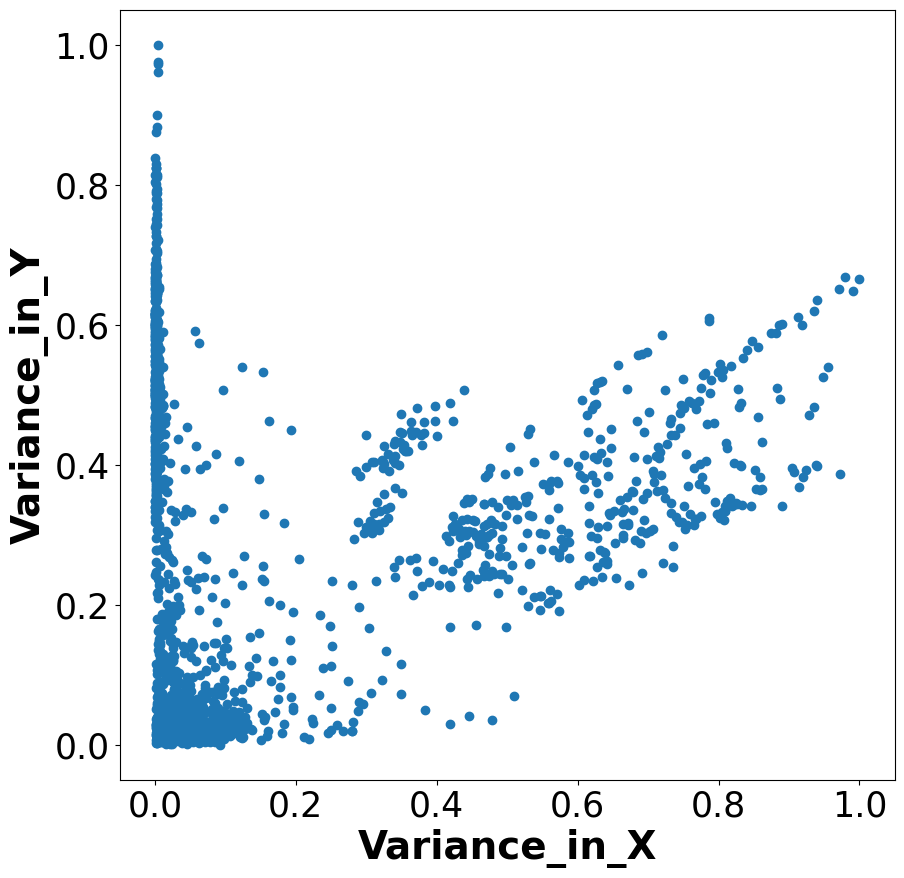}}
\subfloat[After training] {\includegraphics[width=.48\linewidth]{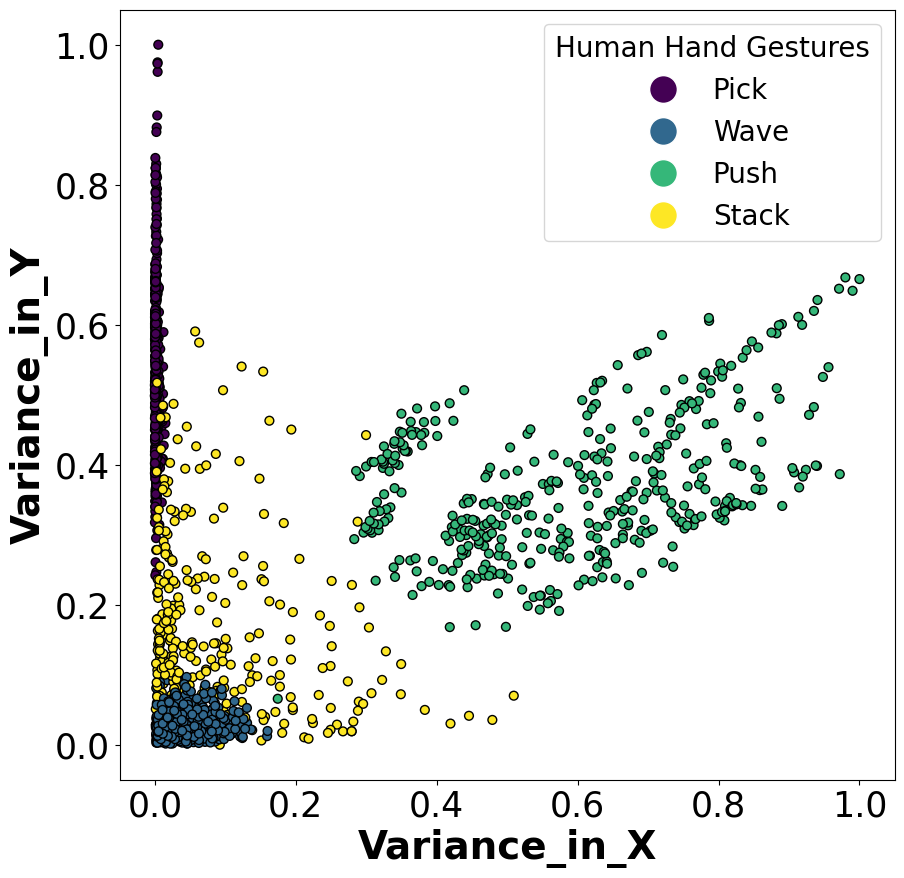}}
\caption{Model training on gesture data for four tasks.(a) Before training: the gesture data were not segregated in accurate clusters. The features (variance in x and y) can not distinguish different gesture types. (b) After training: the features now aligned each gesture data in appropriate clusters.} 
\label{fig: train}
\end{figure}

\begin{figure}
    \centering
    \begin{subfigure}{0.23\textwidth}
        \centering
        \includegraphics[width=\linewidth]{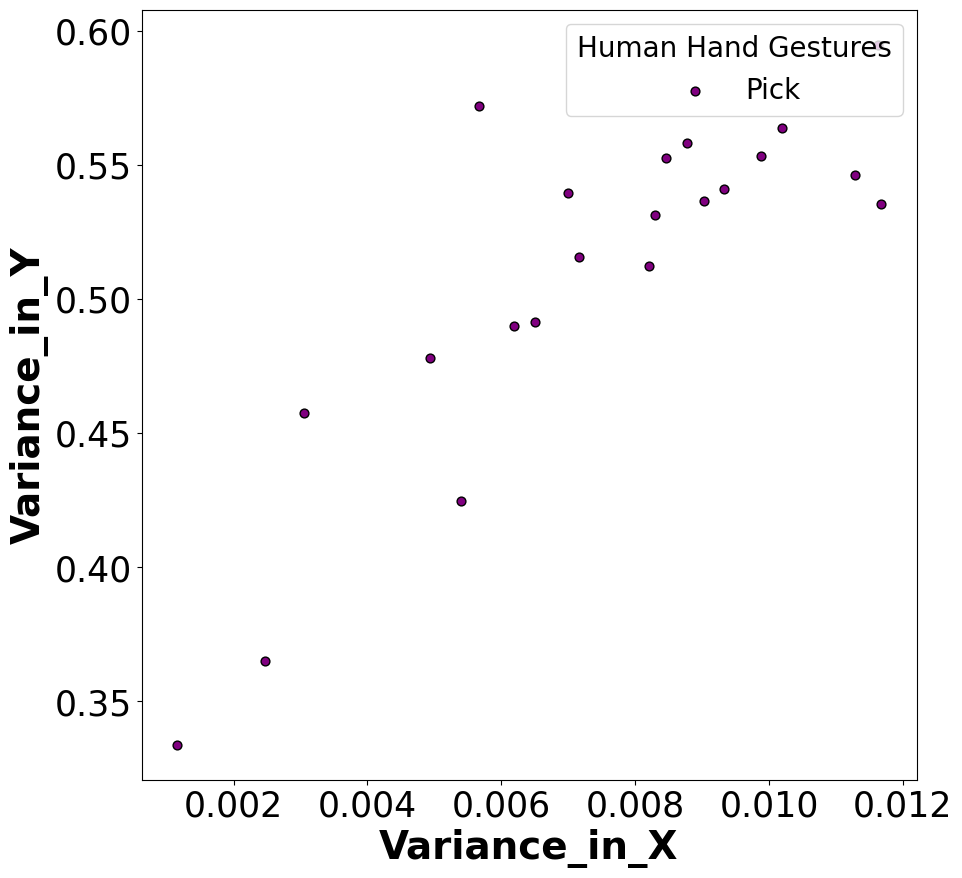}
        \caption{Pick}
        \label{fig:sub1}
    \end{subfigure}
    \hfill
    \begin{subfigure}{0.23\textwidth}
        \centering
        \includegraphics[width=\linewidth]{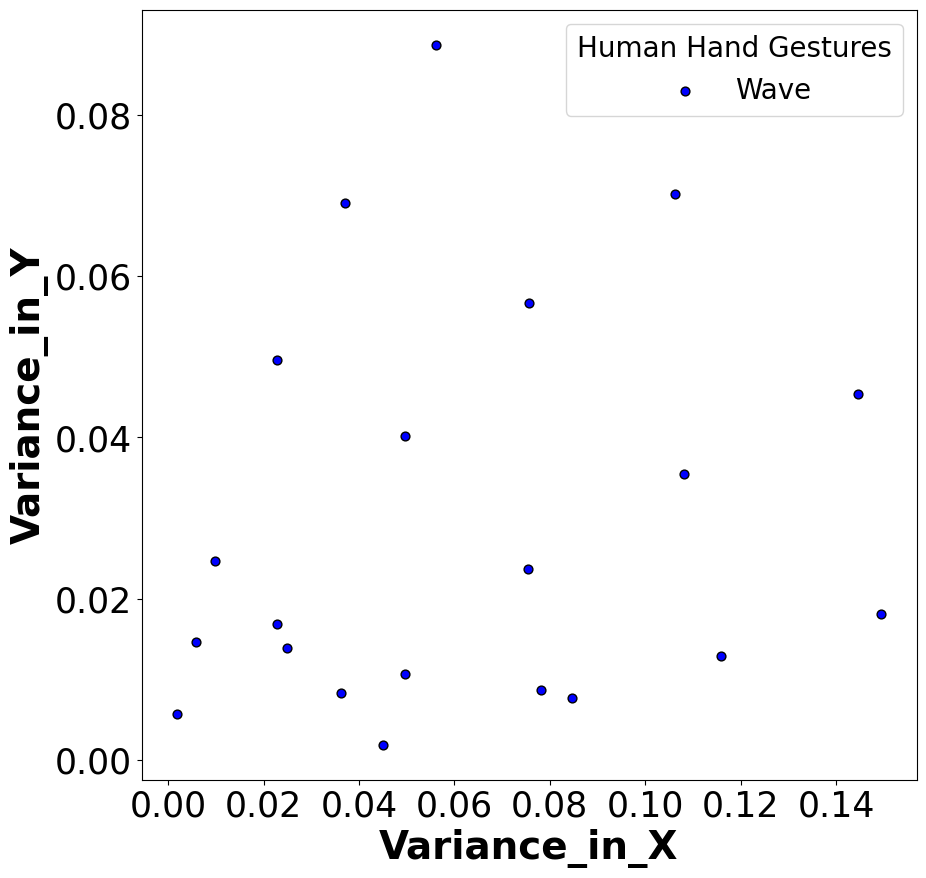}
        \caption{Wave}
        \label{fig:sub2}
    \end{subfigure}

    \medskip

    \begin{subfigure}{0.23\textwidth}
        \centering
        \includegraphics[width=\linewidth]{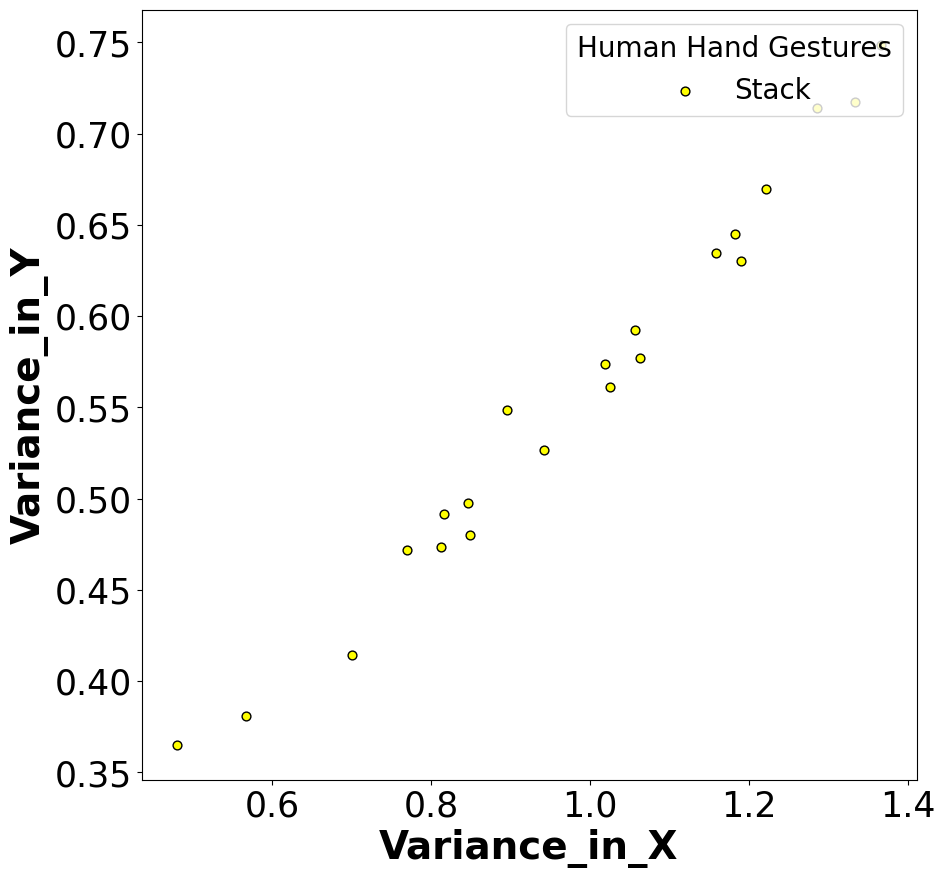}
        \caption{Stack}
        \label{fig:sub3}
    \end{subfigure}
    \hfill
    \begin{subfigure}{0.23\textwidth}
        \centering
        \includegraphics[width=\linewidth]{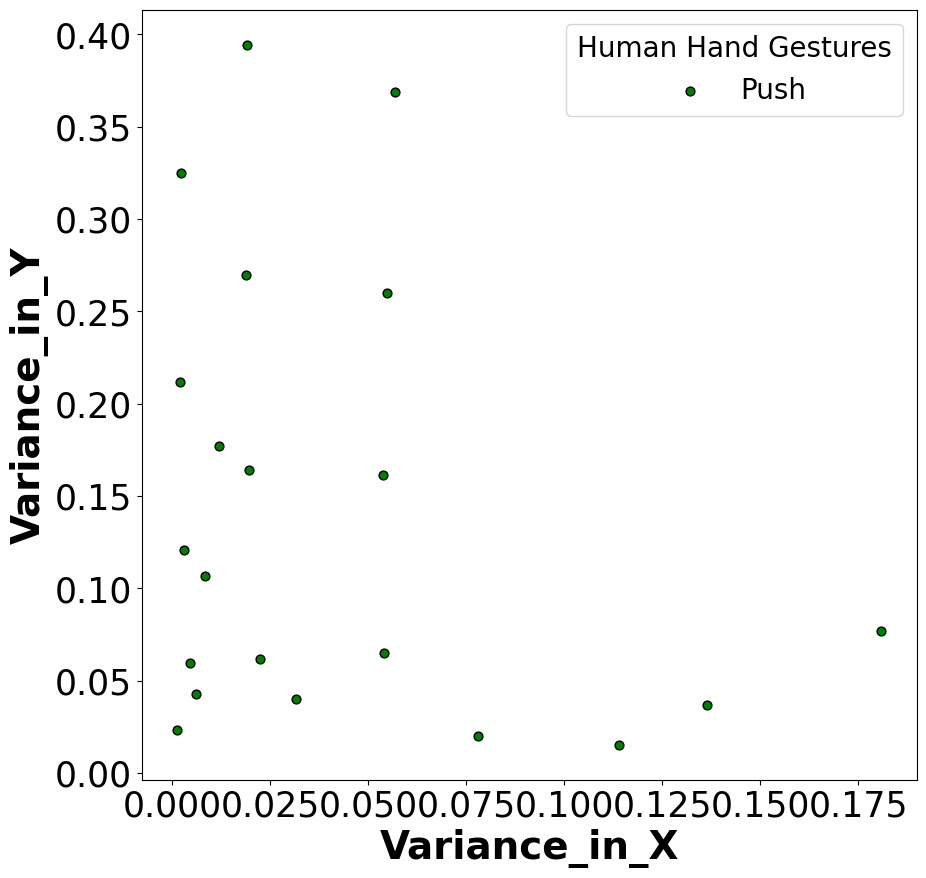}
        \caption{Push}
        \label{fig:sub4}
    \end{subfigure}

    \caption{Model testing on real-time gestures of four tasks. (a) variance largely in $y$ than $x$ direction, (b) variance distributed in both $x-y$ directions, (c) variance follows a uniform pattern in $x-y$ directions, and (d) variance largely in $y$ than $x$ direction.}
    \label{fig: test}
\end{figure}

The total $80\,(20\times 4)$ videos of $5$ seconds each were taken for four distinct hand gestures. 
Each video is converted into 150 frames.
The variance of each landmark point across all frames of a single video was then computed, resulting in a collection of $21\times3$ features in a  CSV file. The dimension is $21\times3$ because of 21 landmark points in each frame and each of them has 3 features($\sigma_x, \sigma_y, \sigma_z$). A total of $1680\times3$ features are there in a CSV file for the $80$ videos (shown in Fig. \ref{fig: train} (a)). The CSV file containing variance features were employed to train our model. The resulting model shown in Fig. \ref{fig: train} (b) was saved for subsequent testing. In the testing phase, the trained model was evaluated using real-time gesture videos. The robot manipulator accurately executed each task based on the identified gestures shown in Fig. \ref{fig: test}. 
To assess the efficacy of our proposed model, we employed the Silhouette score as a metric which varies from -1 to 1. 

Table \ref{training} represents the scores for identifying gestures from videos using the trained model. Table \ref{testing} demonstrates the model Silhouette score during testing. 
The scores of this methodology are found to be better when compared to that in the \cite{Maharani2018ISCAIE}.

\begin{table}[h!]
\caption{Model training results for hand gesture identification.}
\centering
\begin{tabular}{|c|c|c|c|c|}
\hline
\makecell{Gesture\\actions} & \makecell{Number\\ of videos} & \makecell{Detection time \\ (seconds/frame)} & \makecell{Silhouette\\Score} & \makecell{Silhouette\\Score \cite{Maharani2018ISCAIE}} \\
\hline 
Pick & 20 &  0.33 & \multirow[c]{4}{*} {0.6244} & \multirow{4}{*}{0.6081} \\ \cline {1-3} 
Push & 20 & 0.33 &  & \\ \cline{1-3} 
Stack & 20 &  0.33 &  & \\  \cline{1-3} 
Wave & 20 &  0.33 &  & \\
\hline
\end{tabular}
\label{training}
\end{table}

\begin{table}[h!]
\caption{Model testing results for hand gesture identification.}
\centering
\begin{tabular}{|c|c|c|c|c|}
\hline
\makecell{Gesture\\actions} & \makecell{Number\\ of videos} & \makecell{Detection time \\ (seconds/frame)} & \makecell{Silhouette\\Score} & \makecell{Silhouette\\Score \cite{Maharani2018ISCAIE}} \\
\hline
Pick & 2 & 0.33 & \multirow{4}{*}{0.6337} & \multirow{4}{*}{0.6101} \\ \cline {1-3}
Push & 2 & 0.33 &  &  \\ \cline {1-3}
Stack & 2 & 0.33 &  &  \\ \cline {1-3}
Wave & 2 & 0.33 &  &  \\
\hline
\end{tabular}
\label{testing}
\end{table}



We also conducted real-time experiments by showing gestures from $50$ different people related to the $4$ tasks. It achieved accuracy of $94-96\%$ (i.e., out of $50$ tests, the model accurately identified gestures $48$ times).

\section{\textbf{Conclusion}}

This paper proposes a methodology for recognizing different hand gestures with dynamic variations of each in real-time. Prior works that use deep learning models require large training time and a huge amount of data, drastically affecting the identification rate. This approach is based on classical machine learning algorithms that require only a modest computational cost and accurate recognition. The high accuracy leads to the correct robot task execution, thus, promising better human-robot interaction.

The future work will include the development of models that can handle challenges such as: a) the combination of movements in each gesture, and d) two-hand gestures.

\section*{Acknowledgment}
We are thankful to the Technology Innovation Hub(TIH) for Cobotics, IIT Delhi
for funding this work under the project titled
“Healthcare Robotics” reference number IITM/IHFC-IIT DELHI/LB/370.

\bibliographystyle{IEEEtran}
\bibliography{ref}

\section*{Appendix}
\textit{Proof.} The gesture model parameters (mean, co-variance, and mixing weight coefficient) are updated such that the model converges to accurate gesture distributions.
\begin{enumerate}
\item \textit{Mean update:} 

The optimal convergence of the mean parameter for each distribution can be obtained from \eqref{eq7} as

\begin{equation}\label{eq13}
\frac{\partial L}{\partial \mu_k}=\sum_{n=1}^{N} \frac{\partial \log P(X_n\mid \theta)}{\partial \mu_k}=0^{\top}.
\end{equation}

From (\ref{eq13}), we have
\begin{equation}\label{eq15}
 \frac{\partial \log P(X_n\mid \theta)}{\partial \mu_k}=\frac{1}{P(X_n\mid \theta)}\frac{\partial P(X_n\mid \theta)}{\partial \mu_k}.
\end{equation}

Since each gesture follows a normal distribution,

\begin{equation}\label{eq16}
\frac{\partial P(X_n\mid \theta)}{\partial \mu_k}=\sum_{k=1}^{K} \pi_k \frac{\partial \mathcal{N}(X_n\mid \mu_k, \Sigma_k)}{\partial \mu_k}.
\end{equation}
For $k^{th}$ gesture \eqref{eq16} yields
\begin{equation}\label{eq17}
 \frac{\partial P(X_n\mid \theta)}{\partial \mu_k}=\pi_k\frac{\partial \mathcal{N}(X_n\mid \mu_k, \Sigma_k)}{\partial \mu_k}. 
\end{equation}
The gesture distribution (\ref{eq17}) is further simplified as
\begin{equation}\label{eq18}
 \frac{\partial P(X_n\mid \theta)}{\partial \mu_k}=\pi_k (X_n-\mu_k)^T {\Sigma_k}^{-1}\mathcal{N}(X_n\mid \mu_k, \Sigma_k).  
\end{equation}
On using \eqref{eq18} in (\ref{eq13}), yields 

\begin{align}\label{eq19}
 \frac{\partial L}{\partial \mu_k}&= \sum_{n=1}^{N}{\Sigma_k}^{-1} (X_n-\mu_k)^\top \times \notag \\ &\frac{\pi_k\mathcal{N}(X_n\mid \mu_k,\Sigma_k)}{\sum_{J=1}^{K}\pi_j\mathcal{N}(X_n\mid \mu_j,\Sigma_j)}
 =\mathbf{0^\top},      
\end{align}
where $r_{nk}=\frac{\pi_k\mathcal{N}(X_n\mid \mu_k, \Sigma_k)}{\sum_{J=1}^{K}\pi_j\mathcal{N}(X_n\mid \mu_J, \Sigma_J)}$ is the gesture responsibility matrix. Hence,

\begin{equation}\label{eq21}
\sum_{n=1}^{N} r_{nk} (X_n-\mu_k)^\top{\Sigma_k}^{-1}=\mathbf{0^\top},  
\end{equation}
where 

$\Sigma_k^{-1}$ is non-zero. This leads to
\begin{align}\label{eq22}
\sum_{n=1}^{N} r_{nk} X_n =\sum_{n=1}^{N} r_{nk}{\mu_k}^{new}.
\end{align}
Let $N_k=\sum_{n=1}^{N} r_{nk}$. Therefore, the updated mean ($\mu_k$) for a particular gesture distribution is
\begin{align}\label{eq23}
{\mu_k}^{new}=\frac{\sum_{n=1}^{N} r_{nk} X_n}{N_k}.
\end{align} 
\item \textit{Co-variance update:}
The optimal convergence of the co-variance parameter for each distribution can be obtained from \eqref{eq7} as
\begin{equation}\label{eq24}
\frac{\partial L}{\partial \Sigma_k}=\sum_{n=1}^{N} \frac{\partial \log P(X_n\mid \theta)}{\partial \Sigma_k}=0^\top.
\end{equation}

The optimal value of $\Sigma_k$ from \eqref{eq24} can be computed as

\begin{align}\label{eq30}
\frac{\partial \textit{L}}{\partial \Sigma_k} =&-\frac{1}{2}{\Sigma_k}^{-1} N_k +\bigg [\frac{1}{2}{\Sigma_k}^{-1}\times \notag\\ 
&(\sum_{n=1}^{N}r_{nk}(X_n-\mu_k)(X_n-\mu_k)^\top{\Sigma_k}^{-1})\bigg ]
=0^\top.
\end{align}

From \eqref{eq30}, we have

\begin{equation}\label{eq32}
N_k=\sum_{n=1}^{N}R_{nk}(X_n-\mu_k)(X_n-\mu_k)^\top{\Sigma_k}^{-1}.
\end{equation}
Hence, the updated co-variance parameter for a particular gesture is
\begin{equation}\label{eq33}
{\Sigma_k}^{new}=\frac{\sum_{n=1}^{N}R_{nk}(X_n-\mu_k)(X_n-\mu_k)^\top}{N_k}.
\end{equation} 
\item \textit{Mixing weight coefficient update:} 

The optimal convergence of the mixing weight coefficient parameter for each distribution can be obtained from \eqref{eq10} as
\begin{equation}\label{eq34}
\frac{\partial \mathcal{L}}{\partial \pi_k}= 0^\top, \,\,\text{and}\,\,  \frac{\partial \mathcal{L}}{\partial \lambda}= 0^\top.
\end{equation}
The first condition from (\ref{eq34}) yields
\begin{equation}\label{eq35}
\frac{\partial \mathcal{L}}{\partial \pi_k}=\sum_{n=1}^{N} \frac{\mathcal{N}(X_n\mid \mu_k, \Sigma_k)}{\sum_{j=1}^{K}\pi_j \mathcal{N}(X_n\mid \mu_j, \Sigma_j)}+\lambda.
\end{equation}

On further simplifying \eqref{eq35}, 
we have
\begin{equation}\label{eq37}
 \frac{\partial \mathcal{L}}{\partial \pi_k}=\frac{N_k}{\pi_k}+\lambda,  
\end{equation}
where $N_k=\sum_{n=1}^{N} \frac{\pi_k\mathcal{N}(X_n\mid \mu_k, \Sigma_k)}{\sum_{j=1}^{K}\pi_j \mathcal{N}(X_n\mid \mu_j, \Sigma_j)}$.
The optimality condition from \eqref{eq34} in \eqref{eq37} provides
\begin{equation}\label{eq38}
\pi_k=\frac{-N_k}{\lambda}.
\end{equation}
The other condition in \eqref{eq34} is
\begin{equation}\label{eq39}
\frac{\partial \mathcal{L}}{\partial \lambda}= \sum_{k=1}^{K}\pi_k -1 =0^\top. 
\end{equation}
On using $\pi_k$ from (\ref{eq38}) in \eqref{eq39}, we have
\begin{equation}\label{eq40}
-\sum_{k=1}^{K}\frac{N_k}{\lambda}=1 \,\Rightarrow\,\lambda=-N.
\end{equation}
The updated mixing weight coefficient parameter for a gesture distribution using \eqref{eq38} and \eqref{eq40} is
\begin{equation}\label{eq41}
{\pi_k}^{new}=\frac{N_k}{N}.
\end{equation}
\end{enumerate}
During training, all three model parameters are updated optimally each time using \eqref{eq23}, \eqref{eq33}, and \eqref{eq41}. These updated parameters gradually render each gesture distribution to accurate and optimal convergence.

\end{document}